\documentclass[review]{iise}

\usepackage{multirow}
\usepackage{subcaption}

\conference{Proceedings of the IISE Annual Conference \& Expo 2023 \\
K. Babski-Reeves, B. Eksioglu, D. Hampton, eds.}

\title{\titlesize  Multi model LSTM architecture for Track Association based on Automatic Identification System Data }

\author{
Md Asif Bin Syed \&
Imtiaz Ahmed \\ West Virginia University}
\authorlist{Syed and Ahmed}
\abstractID{3649}

\begin{document}
\maketitle

\begin{abstract}

{\small 

For decades, track association has been a challenging problem in marine surveillance, which involves the identification and association of vessel observations over time. However, the Automatic Identification System (AIS) has provided a new opportunity for researchers to tackle this problem by offering a large database of dynamic and geo-spatial information of marine vessels. With the availability of such large databases, researchers can now develop sophisticated models and algorithms that leverage the increased availability of data to address the track association challenge effectively. Furthermore, with the advent of deep learning, track association can now be approached as a data-intensive problem. In this study, we propose a Long Short-Term Memory (LSTM) based multi-model framework for track association. LSTM is a recurrent neural network architecture that is capable of processing multivariate temporal data collected over time in a sequential manner, enabling it to predict current vessel locations from historical observations. Based on these predictions, a geodesic distance based similarity metric is then utilized to associate the unclassified observations to their true tracks (vessels). We evaluate the performance of our approach using standard performance metrics, such as precision, recall, and F1 score, which provide a comprehensive summary of the accuracy of the proposed framework.}
\end{abstract}

\section*{Keywords}
Track Association, Marine Surveillance, Long Short Term Memory (LSTM), Automatic Identification System (AIS), Deep Learning

\section{Introduction}

In contemporary times, technology has advanced to the point where multiple moving objects can be monitored in real-time. However, effectively tracking marine vessels using vast amounts of spatiotemporal data which is collected and gathered in an information system is still challenging. The process of linking unlabeled moving objects to their accurate tracks is referred to as track association and holds significant importance in marine surveillance and national security. The aim of this study is to address track association specifically for Automatic Identification System (AIS) measurements that have already been collected and cleaned for further use.

The AIS is a technology used to track the movements of vessels in real time. It allows ships to exchange information with each other and with land-based receivers using Very High Frequency (VHF) radio signals \cite{Lee}. The information transmitted includes the ship's identity, position, course, and speed, as well as other relevant information \cite{Nguyen}. The potential of this data is vast, with numerous applications in maritime security, environmental monitoring, and traffic management. To harness the full potential of AIS data, efficient tools, and models are needed to extract and analyze relevant information from these streams. In recent years, there have been significant efforts to develop various models from AIS dataset, including track association algorithms, anomaly detection algorithms, clustering, and classification methods. These models can help detect suspicious vessels, predict traffic patterns, and optimize port operations. For instance, they can be used to identify unusual vessel behavior, such as vessels deviating from their usual routes, speeding, or entering restricted areas.

Associating ships with their actual tracks has been a difficult task for many years. Ships exhibit unique behaviors such as slowing down near ports, sudden turns, and unexpected stops, which makes it challenging to track them accurately. Ships may also stop transmitting signals, which can make it difficult to correlate nodes to the correct lane after a long period of no communication \cite{Ahmed}. Moreover, overlapping of the tracks from several vessels make it more challenging for existing algorithms and models to give a better prediction. 

Multiple Object Tracking (MOT) approaches are widely used for track association in marine security and surveillance, utilizing information from multiple sensors such as radar and sonar \cite{Jim}. However, tracking multiple objects presents several challenges, including dealing with an unknown number of objects, sudden appearance and disappearance of objects within the tracking boundary, and uncertain object state \cite{pang}. Sequential tracking algorithms, such as Global nearest neighbor (GNN) and Joint Probabilistic Data Association (JPDA), are also commonly used which are ideally based on the Kalman filtering approach. However, Kalman filtering is mostly limited in handling linear movement of objects. For the non linear movement, the Kalman filtering approach does not perform well \cite{Ahmed}. Physics-based models are also widely used to describe ship motion using mathematical equations and physical laws. These models are useful in the development of simulation systems and training navigation systems. In particular, heuristic and spatio-temporal approaches are two examples of physics-based models that were used to address the track association problem \cite{Ahmed}. However, physics based model uses last known location only to predict the next location and ignores the historical pattern stored in the past observations completely. It can lead to poor performance specially when the last known location is absent or corrupted in some way. 

Track association is a multivariate time series problem that requires consideration of both temporal and spatial components present in the data. The spatio-temporal features in the AIS dataset, such as sequential Time-stamps, Longitude, and Latitude, make it well-suited for deep learning approaches that can process spatio-temporal data. Furthermore, deep learning methods such as (DBSCAN)-based long short-term memory (DLSTM) \cite{Yang},CNN-LSTM \cite{cnnlstm} and Variational recurrent autoencoder \cite{Murray} have good track records in addressing similar problems.Therefore, to address the challenge of processing long sequence of spatio-temporal data, in this work, we develop a multi-model framework based on Long Short-Term Memory (LSTM) networks.  The predictions collected from the LSTM models will be passed through a similarity metric to decide on the final association. Overall, this study aims to address the gap of considering the sequence of spatio-temporal data for the track association problem. 

The rest of the paper is organized as follows. Section 2 presents our problem statement. Section 3 presents the methodology from data preparation to model design and training. Section 3 also presents our experimental results. Finally, Section 4 concludes the paper.

\section {Problem Statement}

\begin{figure}[htb]
	\centering
	\includegraphics[width=4in]{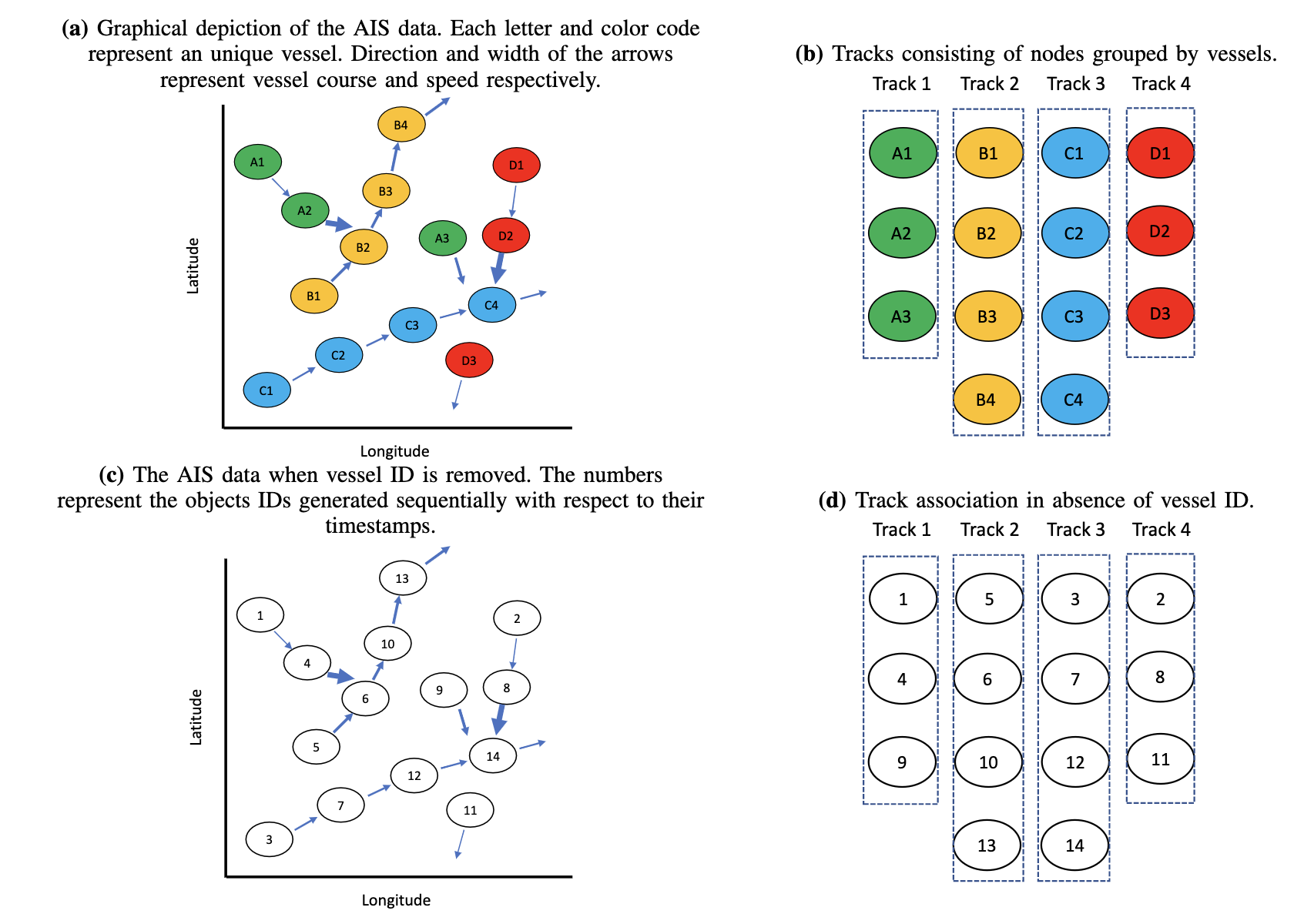}
	\caption{Demonstration of the Association of vessels to their actual tracks \cite{Ahmed}}\label{figure1}
\end{figure}

The AIS dataset was procured from the National Automatic Identification System, maintained by the Coast Guard department of a specific country. Upon receipt of a signal in the AIS system, a unique object ID is assigned to the incoming data, which is then subjected to validation by a monitoring officer to determine if the vessel ID is already present in the existing data. The Figure \ref{figure1}(a) and \ref{figure1}(b) below explains how it has been done. 
If anyhow the vessel information is missing in some timestamps (as shown in Figure \ref{figure1}(c)), then the nodes are instead numbers based on the time they are generated. Mathematically, the problem of track association can be formulated as follows:

Let $m_t\left(u, I_t^k, I_t^s\right)$ to represent an AIS message transmitted at a specific time $t$ where the ship's unique identifier is denoted by  $u$ , and  $I_t^k$ and $I_t^s$ represent the kinematic and static information of the ship at time $t$. A track is defined as a sequence of AIS messages in chronological order broadcast by the same ship shown in Equation \ref{eqn1}\cite{jun}:

\begin{equation}
 T(u)=\left\{m_{t_1}\left(u, I_{t_1}^k, I_{t_1}^s\right), \ldots, m_{t_n}\left(u, I_{t_n}^k, I_{t_n}^s\right) \mid \forall i<j \quad t_i<t_j\right\} 
 \label{eqn1}
\end{equation}

The goal is finding a match between the unknown tracks and the existing tracks as shown in Figure \ref{figure1}(d). The mathematical relationship is shown below in Equation \ref{eqn2} \cite{jun}:

\begin{equation}
\phi\left(m\left(u^{\prime}\right)\right)= \begin{cases}T(u) & m\left(u^{\prime}\right) \text { associated with track } T(u) \\ \varnothing & \text { message forms a new track }\end{cases}
\label{eqn2}
\end{equation}
In the process of matching AIS messages to tracks, it is possible for multiple messages to be associated with a single track, as a single ship may transmit multiple AIS messages. However, it is important to note that each message can only be linked to one ship and cannot be associated with more than one \cite{jun}.

The database utilized in this study was formatted as a CSV file, comprised of comma separated values. A subset of five vessels was selected from the database for the purpose of developing the algorithm, wherein the time intervals between the signals were not even. Upon receipt, the data was transmitted to both regional and national database centers, though the frequency of transmission to the regional center was not uniform. The  irregular time intervals between signals posed a challenge to traditional algorithms used in track association. To evaluate the performance of the algorithms, approximately 2700 data points were trained and 540 vessel IDs (VIDs) were chosen for testing the accuracy of the algorithm. The VID information for these 540 testing data points was deliberately omitted to assess the accuracy of the competing algorithms.

\begin{table}[!ht]
    \raggedleft
    \caption{AIS message stored in  CSV file}
    \begin{tabular}{|c|cccccc|}
    \hline
        \textbf{OBJECT\_ID} & \textbf{VID} & \textbf{SEQUENCE\_DTTM} & \textbf{LAT} & \textbf{LON} & \textbf{SPEED} & \textbf{COURSE} \\ \hline
        1 & 10807db4 & 2020-02-29T22:00:01Z & 37.85671667 & 23.53735 & 0 & 0 \\ 
        2 & 203d4b0c & 2020-02-29T22:00:01Z & 37.9483 & 23.64101667 & 0 & 349.9 \\ 
        3 & 50ee2bf4 & 2020-02-29T22:00:01Z & 37.93902333 & 23.66884833 & 0 & 228.3 \\ 
        4 & 8b998a42 & 2020-02-29T22:00:01Z & 37.93884 & 23.66863333 & 0 & 0.1 \\ 
        5 & 3265e660 & 2020-02-29T22:00:02Z & 37.93147167 & 23.68042667 & 0 & 170.1 \\ \hline
    \end{tabular}
    \label{Table1}
\end{table}

The dataset encompasses seven variables, including object ID, vessel ID, timestamp (comprising both date and time), latitude (expressed in degrees), longitude (expressed in degrees), speed (represented in tenths of knots), and course of direction (represented in tenths of degrees) as presented in (Table \ref{Table1}). 

\section{Methodology}

This section outlines the methodology adopted in our study to address the track association problem. Firstly, Section 3.1 details the data preparation process necessary for feeding our model. Secondly, Section 3.2 provides an in-depth description of the Long Short-Term Memory (LSTM) model architecture that has been trained for predicting the longitude and latitude of subsequent points. The following subsection discusses the assignment of tracks using the Haversine method which is used to compare the predicted output to the actual output. The detailed methodology has been illustrated in Figure \ref{fig1}.

\subsection{Data Preparation}

In this research study, we encountered several challenges when dealing with the large dataset of vessels. Firstly, incorporating a large number of vessels into the algorithms was a time-consuming and complex process. Additionally, the temporal length of the tracks in the dataset was hugely varied, making it difficult to find a suitable common time step. To address these issues, we established a threshold of 500 data points for any vessel to ensure sufficient data for training the model. For this experiment, we included the samples of $Z$ vessels out of all vessels, where $Z$ =5. The sampled tracks can be considered as an irregular time series since the delay between subsequent AIS messages from a single ship can vary greatly over time, resulting in missing data points. These missing data points can disrupt the model's learning process, especially since no pattern was identified in the timestamps of the dataset. 

To address this issue, we employed interpolation of the data and \textit{Re-sampling technique} for every five-seconds interval. This approach allowed us to address the missing data points and create a more regular time series, which could be fed into the model. By using this re-sampled dataset, we were able to overcome the challenges of irregular temporal lengths and missing data points, thereby improving the accuracy and efficiency of our model. For our study, we decided not to include the timestamps as an input feature because we transformed our dataset into an evenly spaced time-series model through the re-sampling technique. To train our model, we employed the Data Scaling method, specifically the min-max scaler, to normalize our input feature variables. The next section elaborates the design of our LSTM model which uses these pre-processed data as input.

\subsection{Longitude and Latitude Prediction using LSTM}

\begin{figure}[htb]
	\centering
	\includegraphics[width=5in]{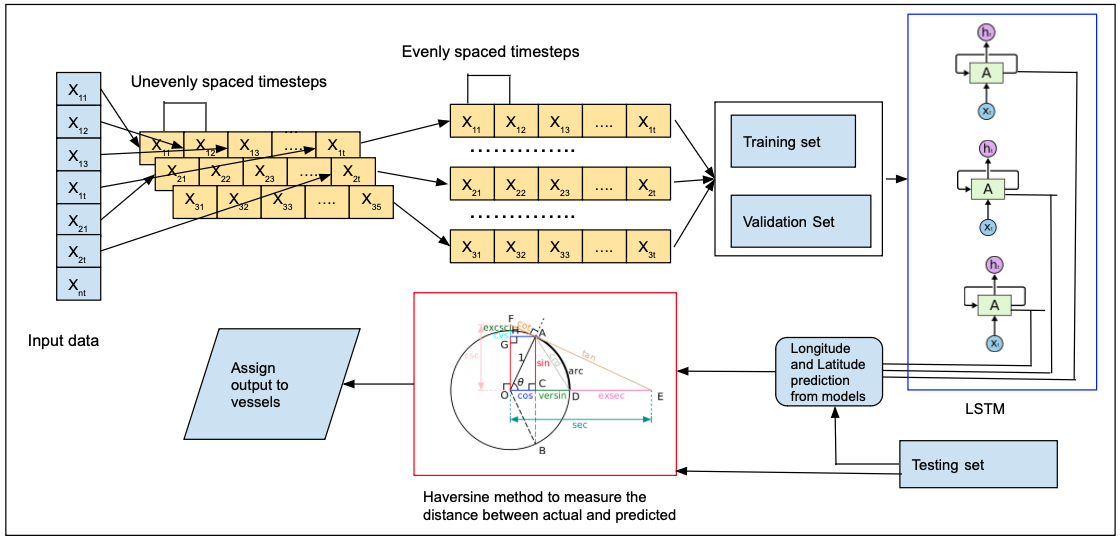}
	\caption{Detailed methodological framework for our proposed multi-model LSTM algorithm}\label{fig1}
\end{figure}

We developed an LSTM model to capture the temporal relationship among the AIS messages of a track. LSTM is a form of recurrent neural network which is widely used  for processing time series data. Our approach involved training each vessel using a separate model in a multi-model fashion. To train the LSTM (shown in the blue outlined box in Figure \ref{fig1}) models in our machine, we divided the available AIS messages into training and testing sets. Let $w$ be the number of testing data, then, from the original dataset $X=\left(x_1, x_2, \ldots, x_{\mathrm{n}}\right)$ of size $\mathrm{n} \times \mathrm{k}$, the training sequences $\left\{x_1, x_2, \ldots, x_{\mathrm{n}-w}\right\}$ and $\left\{y_1, y_2, \ldots, y_{n-w}\right\}$ will be created, where $x_t \in R^{1 x k}$ is the input sequence and $y_t \in R$ is the output data at time $\mathrm{t}$. Here $\mathrm{k}$ and $n$ are the number of features and the total number of observations respectively. To incorporate the required dimension of LSTM architecture, input sequence $x_t$ will be created by taking $m$ continuous sequence $\mathrm{x}_{\mathrm{t}}: \mathrm{x}_{\mathrm{t}+\mathrm{m}-1}$ which is a matrix of shape $\mathrm{m} \times \mathrm{k}$ for $t \in\{1$, $2, \ldots, n-m-1\}$. Here $\mathrm{m}$ is the window size. Window size refers to length of sequence used for prediction. If the model is initiated from $t$=1, the input layer takes the $m$ number of timesteps as inputs and predicts the output $y_{m+1}$. The accuracy and validation loss of the output are then calculated and adjusted into the model. In these types of neural networks, the output is presented as an input in the following step. This feature allows the model to decide based on the most recent input data and the most recent output. So, after predicting the output, $y_{m+1}$ is taken into the input and together with the last $m$ number of timesteps from current timestep, the output $y_{m+2}$ is predicted. This cycle is repeated for all the training data. A training stage pipeline was designed to have multiple parallel input sequences for multivariate values as input, instead of flat input structures for multi-output prediction. Four features will be used which are longitude, latitude, speed and direction of the vessel for generating the multivariate input sequence. Since the vessel’s position needs to be predicted, the longitude and latitude values will be generated as output.
From Table \ref{table2}a, the LSTM layer 1 has the output shape of (none,10,32) where the first dimension, denoted by "None", signifies the dynamic batch size in the Keras model. The second dimension "10" refers to the timesteps that have been used for generating the output and the final dimension refers to the number of neurons. The architecture is made of three LSTM layer for each model. For the first LSTM layer, the return sequences is selected as "True" which means all the hidden state output will be connected to the other layer of the LSTM. The "Relu" is used as an activation function for the LSTM. All the hyperparameters used in training the models are listed in Table \ref{table2}(b) The training of the architecture is done in Tensorflow 2.0. The generated training output is then compared to the actual output using the Haversine formula which is discussed in the next subsection. 

\begin{table}[htb]
\begin{subtable}[c]{0.5\textwidth}
\centering
\begin{tabular}{|l|l|l|}
\hline \textbf{Layer name} & \textbf{Output shape} & \textbf{Param} - \\
\hline LSTM layer 1 & (None, 10, 32) & 4864 \\
\hline LSTM layer 2 & (None, 10, 32) & 8320 \\
\hline Dropout & (None, 10, 32) & 0 \\
\hline LSTM layer 3 & (None, 10, 32) & 8320 \\
\hline Dropout & (None, 10, 32) & 0 \\
\hline Dense layer  & (None, 1) & 33 \\
\hline
\end{tabular}
\subcaption{Output shape and trainable parameter of each phase of the model}
\end{subtable}
\begin{subtable}[c]{0.5\textwidth}
\centering
\begin{tabular}{cc}
\hline Hyperparameters & Value \\
\hline Number of LSTM hidden cells & 32 \\
Return sequences for first layer & True\\
Number of skip connections & 2 \\
LSTM activation function & Relu \\
Batch size & 10 \\
Loss function & Mean Squared error \\
Learning rate & $0.0001$ \\
Epochs & 100 \\
\hline
\end{tabular}
\subcaption{Hyperparameters setting of the models}
\end{subtable}
\caption{Detailed parameters and hyperparameters of the LSTM architecture}
\label{table2}
\end{table}

\subsection{Haversine Method-based Data Association}
Once we estimate the ship’s position at the times of new AIS messages, we need to find a suitable formula to calculate the similarity between the actual location and predicted location. One highly precise geodesic approach for computing the distance between two points on a sphere using their respective latitude and longitude is the haversine formula (illustrated in the red outlined box in the Figure \ref{fig1}). This formula, which is a modified version of the spherical law of cosines, is particularly well-suited for calculating small angles and distances due to its use of haversines \cite{Havershine}. By calculating the Haversine distance using the Equation \ref{eqn3}, the deviation between the predicted location and actual location of the current node can be determined.
\begin{equation}
\operatorname{Cdist}\left(p_k, p_n j\right)=2 r \times \arcsin \sqrt{\sin ^2\left(\frac{\varphi_k-\varphi_n j}{2}\right)+\cos \varphi_k \cos \varphi_n \sin ^2\left(\frac{\lambda_k-\lambda_n j}{2}\right)}
\label{eqn3}
\end{equation}
where, $\mathrm{p}_{\mathrm{k}}:=\left(\varphi_k, \lambda_k\right)$ and $\mathrm{p}_n \mathrm{j}:=\left(\varphi_n \mathrm{j}, \lambda_n \mathrm{j}\right)$. here $\mathrm{p}_{\mathrm{k}}$ and $\mathrm{p}_n \mathrm{j}$ refers to two specific position such as predicted and output location. Additionally, $\varphi_k$ and $\varphi_n\mathrm{j}$ are the the longitudes of the positions, whereas $\lambda_k$ and $\lambda_n \mathrm{j}$ are the latitudes of the positions.

\section{Results and Discussion}
To assess the efficacy of our model, we calculated its confusion matrix. The confusion matrix is a representation of the actual classes versus the predicted classes, with the main diagonal of the matrix representing the true positive predictions made by the model. However, the confusion matrix alone is insufficient in quantifying the performance of the model.

\begin{figure}[ht]
  \centering
  \begin{subfigure}[b]{0.4\textwidth}
    \includegraphics[width=\textwidth]{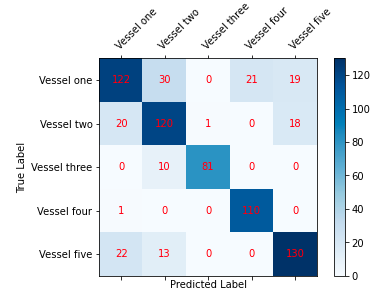}
    \caption{Confusion matrix for the test set}
    \label{fig:image}
  \end{subfigure}
  \qquad
  \begin{subfigure}[b]{0.5\textwidth}
    \centering
    \begin{tabular}{ll|cccc}
\hline
\multicolumn{2}{l}{}  & Precision & Recall & Accuracy & F1 score \\ \hline
\multirow{3}{*} & Vessel One & 0.725 & 0.670 & 0.696& 0.730\\ \cline{2-6}
 & Vessel Two & 0.712 & 0.782 & 0.745 & 0.768\\ \cline{2-6}
 & Vessel Three & 1.000 & 0.890 & 0.942 & 0.964\\ \cline{2-6}
& Vessel Four & 0.840 & 0.990 & 0.909 & 0.992\\ \cline{2-6}
& Vessel Five & 0.826 & 0.760 & 0.792 & 0.804\\ \cline{2-6}
 
\end{tabular}
    \caption{Evaluation metrics results for the model}
    \label{tab:table}
  \end{subfigure}
  \caption{Performance comparison for all the vessels using the proposed model}
  \label{fig:figure}
\end{figure}

To fully gauge the model's performance, we computed several widely recognized evaluation metrics, including: Precision = TP / (TP + FP), Recall = TP / (TP + FN), Accuracy = (TP + TN) / (TP + TN + FP + FN), and F1 Score = 2 * (precision * recall) / (precision + recall) where TP stands for true positive, TN for true negative, FP for false positive, and FN for false negative predictions. 

The evaluation of the model on the test dataset yielded promising results. The confusion matrix and evaluation metrics for the five different vessels are presented in Figure \ref{fig:figure}(b). The table shows that the model was able to correctly predict the class of Vessel Three with 100\% precision, 89\% recall, 94\% accuracy, and an F1 score of 0.964. Vessel Four also had high scores in all metrics except precision. Vessel Five had the third-highest scores in precision, recall, and F1 score, while Vessel One and Vessel Two had relatively lower scores in all metrics. The overlapping for a good length of the data points might be the reason behind lower score for these vessels. These results indicate that the model has performed well in predicting the class of vessels in the test dataset, with high accuracy, precision, recall, and F1 score for most of the vessels. However, it is also evident that the model needs improvement for the prediction of Vessel One and Vessel Two, as they had relatively lower scores in all metrics.

\section{Conclusion}

In summary, we can conclude that mining temporal patterns for future prediction using LSTM architecture is an effective approach for vessel location prediction. When coupled with a geodesic similarity metric, it can successfully associate vessel locations to their true tracks. However, our approach is purely a data-driven approach and thus ignores the underlying physics of the vessel movements. In future, it would be worth investigating the possibility of a physics-based LSTM model where the physics model will be guiding the LSTM predictions.

\end{document}